\pdfoutput=1

\documentclass[11pt,a4paper]{article}
\usepackage{acl2017}
\usepackage{times}
\usepackage{latexsym}

\usepackage{url}
\usepackage{latexsym}
\usepackage{amssymb}
\usepackage{amsmath}
\usepackage{txfonts}
\usepackage{placeins}
\usepackage{rotating}
\usepackage{floatrow}
\usepackage{todonotes}

\aclfinalcopy % Uncomment this line for the final submission
\title{Language Independent Acquisition of Abbreviations}

\author{Michael R. Glass, Md Faisal Mahbub Chowdhury and  Alfio M. Gliozzo \\
  IBM Research \\
  51 Astor Place, New York, NY 10003 \\
  {\tt \{mrglass, mchowdh, gliozzo\}@us.ibm.com} \\}

\date{}

\begin{document}
\maketitle
\begin{abstract}
This paper addresses automatic extraction of abbreviations (encompassing acronyms and initialisms) and corresponding long-form expansions from plain unstructured text. We create and are going to release a multilingual resource for abbreviations and their corresponding expansions, built automatically by exploiting Wikipedia redirect and disambiguation pages, that can be used as a benchmark for evaluation. We address a shortcoming of previous work where only the redirect pages were used, and so every abbreviation had only a single expansion, even though multiple different expansions are possible for many of the abbreviations. We also develop a principled machine learning based approach to scoring expansion candidates using different techniques such as indicators of near synonymy, topical relatedness, and surface similarity. We show improved performance over seven languages, including two with a non-Latin alphabet, relative to strong baselines.
\end{abstract}

\section{Introduction}

An abbreviation is a shortened or contracted form, usually represented by a single token, of a term (comprising a single word or multiple words). Usage of abbreviations is very common in almost all types of text genres. Specifically, technical documents often contain a considerable number of abbreviations. Widespread use of electronic communication through popular textual communication services such as SMS, Tweets, etc, that are generally restricted to a certain maximum number of characters, has driven even more usage of abbreviations.   For example, it was estimated that 10\% or more of the words in a typical SMS message are abbreviated \cite{Crystal2008}. Hence, identifying abbreviations (or short-forms) and mapping them to the appropriate expansions (or long-forms) has applications in query expansion, corpus indexing, terminology extraction and ontology population, among others. 

Like natural language in general, abbreviations can be ambiguous based on context and also because of variations in possible expansions. For example, the short form ``ACE'' can have a large number of expansions including ``accumulated cyclone energy'' and ``American Council on Education''. Additionally ``ace'' may not be a short form in some context, instead referring to a playing card or an exceptional individual. 

In this work, we address the construction of a dictionary to map short-forms to corresponding (one or more) long-form expansions from given plain text. We focus on developing a language independent solution that can be effective across a broad range of domains including financial, medical and legal and a number of different genres, such as news articles, expository text, and yearly reports. We do not address the disambiguation task in this work, only the construction of the resource mapping each short-form to its possible expansions.

%More formally, our task is to construct a mapping from short form to set of possible long forms. This list is scored by confidence.

To be more specific, this work makes the following contributions - 

\begin{itemize}

\item We create and are going to release a multilingual resource for abbreviations and expansions, extracted automatically using Wikipedia redirect and disambiguation pages,  that can be used as a benchmark for evaluation.
%\item A shortcoming of previous work is that only the Wikipedia redirect pages were used, and so every abbreviation had only a single expansion even though multiple different expansions are possible for some of the abbreviations. %We develop a principled machine learning based approach to scoring expansion candidates using different techniques such as indicators of near synonymy, topical relatedness, and surface similarity. 
%\item We remove from the ground truth cases where the short-form and long-form do not co-occur in the same document (in Wikipedia), giving us a denominator for recall that is actually achievable by an extraction system
\item We explore evaluation methodologies for this ground truth, using both an automatic evaluation and a partly manual assessment.
\item We develop a principled machine learning based approach to scoring expansion candidates using different techniques: indicators of near synonymy, topical relatedness, and surface similarity. 
\begin{itemize}
\item We develop an architecture to use word embeddings for assessing semantic similarity between short and long forms.
\item We develop a max-over-alignments model for determining the surface similarity of a short and long form.
\end{itemize}
\item We show improved performance over seven languages, including two with a non-Latin alphabet, relative to strong baselines without exploiting any kind of stop word list.
\end{itemize}

\section{Related Work}

In this section, we briefly describe some of the related work. Most of the previous work for abbreviation detection and corresponding expansion extraction has been focused on acronyms, and in the context of English biomedical text. The study reported by Torii et al  \shortcite{Torii2007} contains a comprehensive survey of these works. 

A common strategy in the literature is to identify occurrences where an abbreviation is explicitly paired with its expansion \cite{Schwartz2003}, \cite{Ehrmann2013}. By far the most frequent way to express this relationship in text is through a pattern involving a parenthetical: \textit{long form (SF)} or \textit{SF (long form)}, where ``SF'' is the short-form or abbreviation and ``long form'' is the expansion.  The widely cited method of Schwartz and Hearst \shortcite{Schwartz2003} identifies these patterns in text and uses a rule-based procedure to determine if the parenthetical is expressing an abbreviation relation, and in the case of the first and more common pattern, \textit{long form (SF)}, the boundary of the long-form. 

Some of the strategies used by this method are adopted by almost all the later work. For example, the candidate long-form must appear in the same sentence as the short-form; the long-form must contain all of the characters of the short-form; the short-form characters must be in order in the long-form; the first letter of the short-form and the long-form must be the same, etc. This method was developed for English biomedical text but, to some extent, is applicable to multiple languages, since it does not require any language specific resource or training.

In general, rule-based approaches find short-forms according to rules based on case and punctuation. Long-form candidates are gathered from a window around short-form mentions, possibly requiring a connecting pattern like parentheses. The short-form and long-form are then matched by rules, usually according to the occurrence and order of the short-form letters in the long-form and often requiring a language specific stop-word list for filtering potential errors in output \cite{Taghva1999}, \cite{Larkey2000}, \cite{Pustejovsky2001}, \cite{Yu2002}. In fact, even many machine learning based systems also use a stop-word list \cite{Okazaki2006}, \cite{Ehrmann2013}.

Terada \shortcite{Terada2004} tabulated the contexts of short-form occurrences and long-form occurrences using both an abbreviation rich and abbreviation poor corpus. The short-forms and long-forms were then scored according to their distributional similarity, by measuring the cosine of their context vectors.

Supervised approaches identify candidate short and long forms according to patterns and filters, then generate features based on the string similarity of the short and long form and the characteristics of the short and long form. Logistic regression or SVMs are used to classify \cite{Chang2002}, \cite{Nadeau2005}. However, these efforts did not improve on unsupervised, rule based approaches, such as Schwartz and Hearst\shortcite{Schwartz2003}.

Noisy supervision has been used to avoid extensive annotation effort \cite{Yeganova2010}. Terms matching certain filters found in parentheses patterns are used as potential short-form/long-form pairs. This dataset is then used to train features for detecting abbreviations.

Okazaki and Ananiadou \shortcite{Okazaki2006} extract long-forms from the contextual sentences (i.e.~sentences where candidate long-form and short-form co-occur) in a similar manner to the term recognition task. 

A common evaluation problem in most of the previous reported result is that they do not consider multiple occurrences of the same \{short-from, long-form\} pair uniquely during scoring. In other words, if a certain pair \{A, B\} has twenty occurrences in text, and if the system is able to detect \textbf{``B''} is a long-form of \textbf{``A''}, the system will have a match score of twenty instead of one. This is problematic because the scoring tends to bias towards pairs with multiple occurrences. The more occurrences a pair has, the more likely it to be extracted by a system regardless of whether it uses co-occurrence statistics, patterns or features (for ML based systems). 

Most of the multi-lingual (not language independent) work generalize rule-based matching methods but require multiple linguistic resources for each target language \cite{Zahariev2004thesis}. Ehrmann et al \shortcite{Ehrmann2013} uses a slightly modified version (by exploiting a multilingual stop-word list) of the Schwartz and Hearst \shortcite{Schwartz2003} method for evaluating performance for seven languages.

\section{Dataset} 
\label{sec:dataset}

A major obstacle to comparing the robustness and effectiveness of different approaches is the lack of large benchmark dataset in a general domain, even for English. There exist some small corpora for English in the biomedical domain, e.g. Medstract gold standard evaluation corpus \cite{Torii2007} and BioText corpus \cite{Schwartz2003}, which are not sufficient to assess the general applicability of an approach. Some available corpora (for English), originally depeveloped for other tasks, do contain a lot of abbreviations (e.g. the TR9856 term relatedness benchmark data \cite{Levy2015}), but it is not possible to use them as they do not identify the type (abbreviation, antonym, etc.) of relation between two related terms. %are for the benchmarking of abbreviation extraction systems. 

Jacquet et al \shortcite{Jacquet2014} proposed a large multilingual dataset where they made use of the Wikipedia redirection pages for generating gold standard data. A shortcoming of this work is -- because of the usage of only the redirect pages for gold standard annotation, every abbreviation has only a single expansion even though multiple different expansions are possible for some of the abbreviations.  This resource was only developed and used for Latin alphabet languages.

%A number of resources have been developed for the evaluation of abbreviation extraction systems.  In the medical domain \todo{give medical domain resource}. Ehrmann et al. \shortcite{ehrmann2013acronym} developed a multilingual resource for Latin alphabet languages using Wikipedia redirect pages.

We have extended the method of mining Wikipedia to gather a multilingual ground truth. By mining disambiguation as well as redirect pages we can gather a ground truth including short-forms with multiple possible long-form expansions. For most languages, the addition of short-form / long-form pairs mined from disambiguation pages more than doubles the amount of ground truth available. Unlike redirect pages, disambiguation pages are indicated using different Mediawiki markup for each language, necessitating the construction of a language-aware system for gathering the ground truth. Abbreviation extraction for non-Latin alphabet languages such as Russian and Japanese can also be evaluated using our resource. %TODO: clarify what we mean by ground truth here, move explanation of pseudo-precision here

%explain what the redirect and disambiguation pages are
We processed the Wikipedia dump using Sweble \cite{sweble} to get the article text from the Mediawiki markup dump and to parse the redirect and disambiguation pages.
Redirect pages serve to send links from one Wikipedia page name to another, typically because the former is a synonym for the later. For example 'URL' redirects to 'Uniform Resource Locator'. Disambiguation pages exist when a term is ambiguous and there exist multiple Wikipedia pages that may be relevant, depending on the intended meaning. The 'URL (disambiguation)' page links to 'Uniform Resource Locator' as well as 'Unrestricted Line Officer' and an artist named 'URL'.

We focus on the seven languages with the most Wikipedia articles: English, German, French, Italian, Spanish, Russian, and Japanese.

\begin{table}[htb!]
\begin{center}
\vspace*{0.6in} %room for rotated column labels
\begin{tabular}{c|r|r|r|r|r|}
\bf Language & 
\begin{rotate}{60} Short-forms \end{rotate} &
%\begin{rotate}{60} Pairs \end{rotate} &
\begin{rotate}{60} Redirects \end{rotate} &
\begin{rotate}{60} Disambig. \end{rotate} &
\begin{rotate}{60} Articles \end{rotate}
 \\ \hline 
%English & 45714 &	69323 & 37157 & 32166 & 4884754 \\ \hline 
%German & 13543 & 29159 & 5284 & 23875 & 2215751 \\ \hline 
%French & 7750 &	11941 & 4084 & 7857 & 1675260 \\ \hline
%Italian & 5300 & 8791 & 1688 &	7103 & 1399494 \\ \hline 
%Spanish & 3752 & 6143 & 2300 & 3843 & 2297591 \\ \hline
%Russian & 2195 & 2886 & 1479 & 1407 & 1854549 \\ \hline
%Japanese & 2084 & 2484 & 918 & 1566 & 1150371 \\ \hline

English & 45714 & 37157 & 32166 & 4.88M \\ \hline 
German & 13543 & 5284 & 23875 & 2.22M \\ \hline 
French & 7750 &	4084 & 7857 & 1.68M \\ \hline
Italian & 5300 & 1688 &	7103 & 1.40M \\ \hline 
Spanish & 3752 & 2300 & 3843 & 2.30M \\ \hline
Russian & 2195 & 1479 & 1407 & 1.85M \\ \hline
Japanese & 2084 & 918 & 1566 & 1.15M \\ \hline
\end{tabular}
\end{center}
\caption{\label{datasetTable} Statistics on the multilingual abbreviation dataset}
\end{table}

Table \ref{datasetTable} gives statistics on the dataset for each language. The column \textit{`Short-forms'} gives the number of unique short-forms. %, while \textit{Pairs} is the number of short-form / long-form pairs. 
\textit{`Redirects'} and \textit{`Disambig.'} break down the ground truth short-form / long-form pairs into those gathered from redirect pages and those gathered from disambiguation pages. \textit{`Articles'} is the number of articles, excluding redirect and disambiguation pages.

Redirects and disambiguation pages contain many substitutable or closely related terms that are not abbreviations. We filtered the extractions to exclude pairs that do not fit broad criteria for an abbreviation.
\begin{itemize}
\item Short-forms are restricted to ten characters or less.
\item For case-sensitive languages, at least half of short-form characters must be uppercase.
\item The long-form must be at least twice as long as the short-form, with at least two tokens.
\item The long-form must not contain the short-form as a substring.
\item 80\% of the characters in the short-form must appear in the long-form.
\end{itemize}

We are aware that this multilingual ground truth is not complete. Firstly, not all abbreviations inside the Wikipedia articles have links for redirects or disambiguation pages. Secondly, although the inclusion criteria mentioned above are very broad, they do not cover every possible case. For example, CA is an abbreviation of California but it will be filtered by one of our criteria (long-form must not contain the short-form as a substring). Nevertheless, we believe that this resource of multilingual ground truth has enough coverage to be used as a benchmark. Furthermore, we show later that the incompleteness of the ground truth is not a problem for ranking different systems.
 
During evaluation of the systems (described later), we filter the extractions of all systems compared using the same criteria, to avoid penalizing methods for producing short-form / long-form pairs that do not fit our criteria for an abbreviation.

\section{Approach}

%TODO: three sources of evidence: how the terms occur in text (sometimes the connection between a short-form and long-form is explicitly stated); their meaning as seen through usage pattern; the surface or string similarity between the two terms.

Our approach operates in two basic stages: candidate generation and scoring. For candidate generation we use two systems, each of which provides, or can be adapted to provide a score. The first system is developed by Schwartz and Hearst \shortcite{Schwartz2003} which is freely available\footnote{http://biotext.berkeley.edu/software.html} and delivers performance close to the state-of-the-art. This system identifies occurrences of the parentheses pattern that indicate a short-form / long-form pair. To convert this output into a scored expansion dictionary we simply gather the occurrence count for each extracted pair as the score. 

The second candidate generation system is a rule based system we developed as part of this work. We will refer to this system as Candidate System 2 later. %We provide a brief description of this system here due to the 8 page space limitation for this submission. 
This system does not use the parentheses pattern and therefore provides a different set of candidates. It produces a prediction score based on the co-occurrence frequency of the pairs in the whole corpus, minimum distance (in number of tokens) between their occurrences in text, and approximate string similarity of the short-form and long-form. A candidate pair is discarded if the prediction score is below a certain threshold.

Candidates are gathered by running the above two extraction systems on the Wikipedia article plain text corpus for each language, excluding of course the redirect and disambiguation pages, since these were used to build the ground truth.

The second stage, scoring, is one of the main contributions of this paper.  Abbreviations are shortened forms with the same semantics as their expansion. We further refine the notion of semantics into \textit{near-synonymy} and \textit{topical relatedness}. Near-synonymy means that one term can be replaced with the other, while generally preserving the meaning of the sentence. In other words, near-synonyms are substitutable terms. Topical relatedness means that the two terms occur in the same sorts of documents, genres, or topics. 

\begin{figure}[htb]
   \centering
   \includegraphics[width=2.3in]{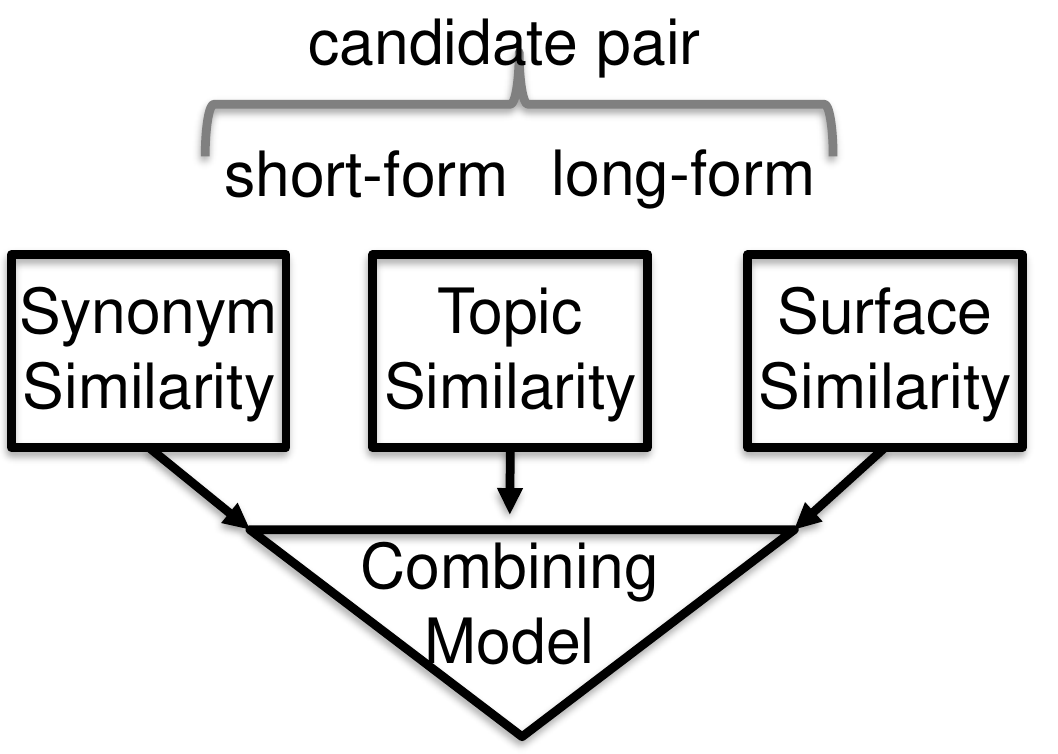}
   \caption{Abbreviation Scoring Architecture}
   \label{architecture}
  \end{figure}

Figure \ref{architecture} gives an overview of the scoring architecture. The three aspects of an abbreviation pair are combined into a single score using a combining model. In our case, we use a simple logistic layer.

\subsection{Semantic Similarity by Word Embeddings} 
\label{sec:semanticFeatures}

To assess the semantic similarity of the short-form and long-form we use word embeddings, specifically word2vec and LSA. The short-form, or abbreviation is given by the single token $\alpha$, while the tokens $T = \tau_1 \tau_2 \tau_3 ...$ constitute the long-form.
A word embedding lookup table $LT_W$ is parameterized by the word vectors $W \in \varmathbb{R}^{d \times |V|}$, for the vocabulary $V$ with a hyper-parameter dimension $d$. 

The lookup table is used to project the tokens into a vector space. In the case of the long-form, the word vectors are pooled by average to obtain a single vector. The two vectors are then compared for cosine similarity, a standard metric for vector space similarity.  We also produce features for the exponentiated similarity, $e^{sim}$, and logit of the re-centered similarity, $\mathbf{logit}((sim+1)/2)$, since these transformations may have a more linear relationship with the log-odds of a correct short-form / long-form pair.

\begin{align*}
\vec{\sigma} & = LT_{W}(\alpha) \\
\vec{\lambda} & = \frac{1}{|T|} \sum_{\tau_i \in T} LT_{W}(\tau_i) \\
sim &= \frac{\vec{\sigma} \cdot \vec{\lambda}}{||\vec{\sigma}|| \cdot ||\vec{\lambda}||}
\end{align*}

If one of the long-form tokens is out-of-vocabulary for the word embedding model, then that token is ignored. If all tokens for either the long-form, or the single token of the short-form is out-of-vocabulary, then the word embedding similarity is absent, and a separate missing feature is output.

\begin{figure}[htb]
   \centering
   \includegraphics[width=2.5in]{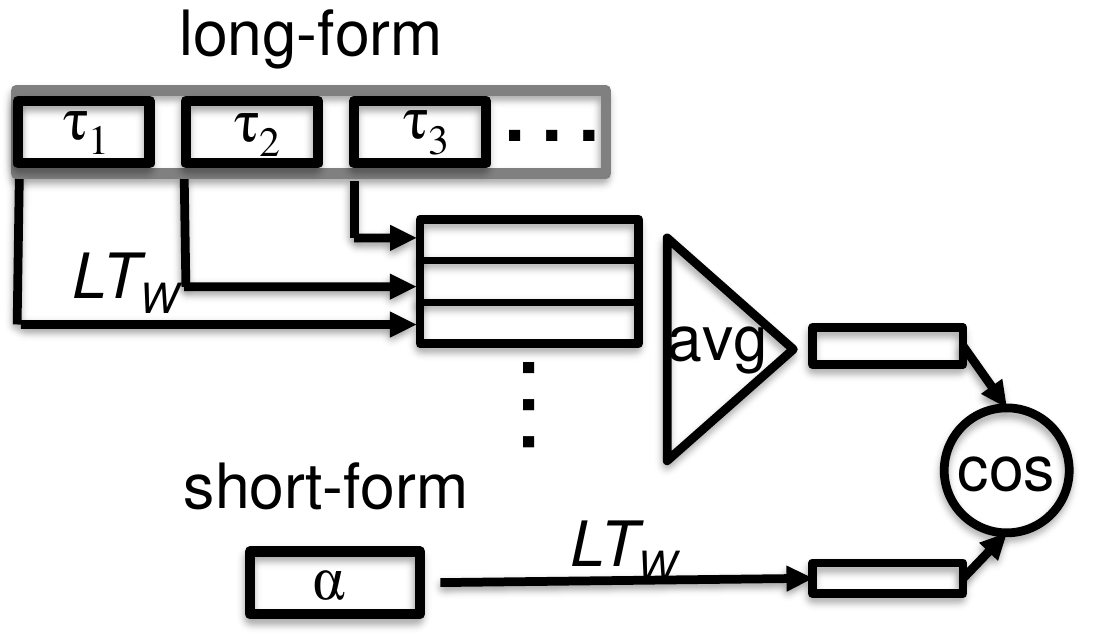}
   \caption{Word Embedding Similarity}
   \label{embeddingSim}
  \end{figure}

Figure \ref{embeddingSim} illustrates the architecture for measuring semantic similarity. This architecture is applied both for near-synonymy and topical relatedness, with appropriate word embeddings $W$ used in the lookup table $LT_W$. 
The word embeddings are trained on the Wikipedia article plain text corpus for each language, again excluding redirect and disambiguation pages.

\subsubsection{Near-Synonymy}

We use the vector space embeddings of the word2vec \cite{word2vec} system to determine near-synonymy (substitutability). The continuous bag-of-words (CBOW) model of word2vec is a natural fit for this goal, since the task for which the vectors are trained is to predict a word given a continuous representation of its bag-of-words context.  The CBOW model averages the context vectors (a learned parameter) for the words in a window around a focus word that is predicted. The dot product of the average context vector with a particular word vector is then interpreted (after applying a logistic) as the probability that the word is present in the context. As a result of training with the objective of minimizing the log-probability loss, words that appear in similar contexts have similar vectors.

This is similar to the method of Terada \shortcite{Terada2004}, but uses neural word embeddings rather than constructing sparse vectors from a simple count of context words.

\subsubsection{Topical Relatedness}

To assess the topical similarity we use the method of Latent Semantic Analysis (LSA) \cite{lsa}. LSA constructs a sparse term-by-document matrix and factorizes it by singular value decomposition (SVD) to obtain dense vector representations of terms, in our case, single tokens. The matrix factorization generalizes the word embeddings, so that two word vectors can be topically similar if they are in documents with similar topics, even if they do not happen to occur in any documents together.

Additionally we include as features raw co-occurrence counts of the full long-forms and short-forms, as well as the frequencies of the individual short or long forms, both in log space. The co-occurrence count can serve as a simpler form of topic model.

\subsection{Surface Similarity by Alignment} \label{sec:alignmentFeatures}

Abbreviations are shortened forms of a longer term and therefore have a mapping from their surface form, to the surface form of the long-form. Previous work has either produced surface features that do not depend on a particular mapping or alignment \cite{Nadeau2005}, or has used a single, pre-defined method of finding the best alignment \cite{Chang2002}. We use a maximum over possible alignments architecture to learn both how short and long forms align, and how well any given short-form / long-form pair can be aligned.

\begin{figure}[htb]
   \centering
   \includegraphics[width=3.0in]{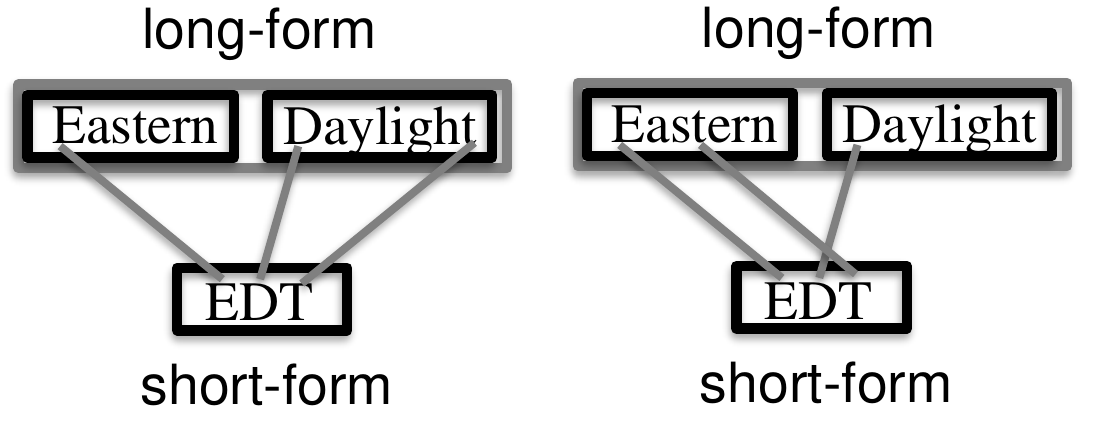}
   \caption{Possible Alignments}
   \label{aligns}
  \end{figure}

For a given short-form / long-form pair we consider a \textit{possible alignment} to be a mapping from each character of the short-form to either 1) a matching (ignoring case) character of the long-form, or 2) a null match. Figure \ref{aligns} shows two possible alignments of the short-form ``EDT'' to the long-form ``Eastern Daylight''. Note though, that all of the alignments between this pair are wrong, since the appropriate expansion is ``Eastern Daylight Time''.

%\begin{minipage}{\columnwidth}
Given a possible alignment, features can be extracted. We consider the following features:
\begin{itemize}
\item Number of short-form characters that match a token-initial character in the long-form.
\item Percentage of short-form characters that match a token-initial character in long-form.
\item Percentage of long-form tokens that have no short-form character match.
\item Percentage of short-form characters mapped to the null match.
\item Number of swaps in the order of the short-form mapped characters. For example, a swap occurs if the first character of the short-form matches to a character in the long-form after the long-form character that the second character of the short-form is mapped to.
\end{itemize}
%\end{minipage}

The percentage features are transformed by exponentiation to better relate their value to a log-odds. 

\begin{figure}[htb]
   \centering
   \includegraphics[width=3.0in]{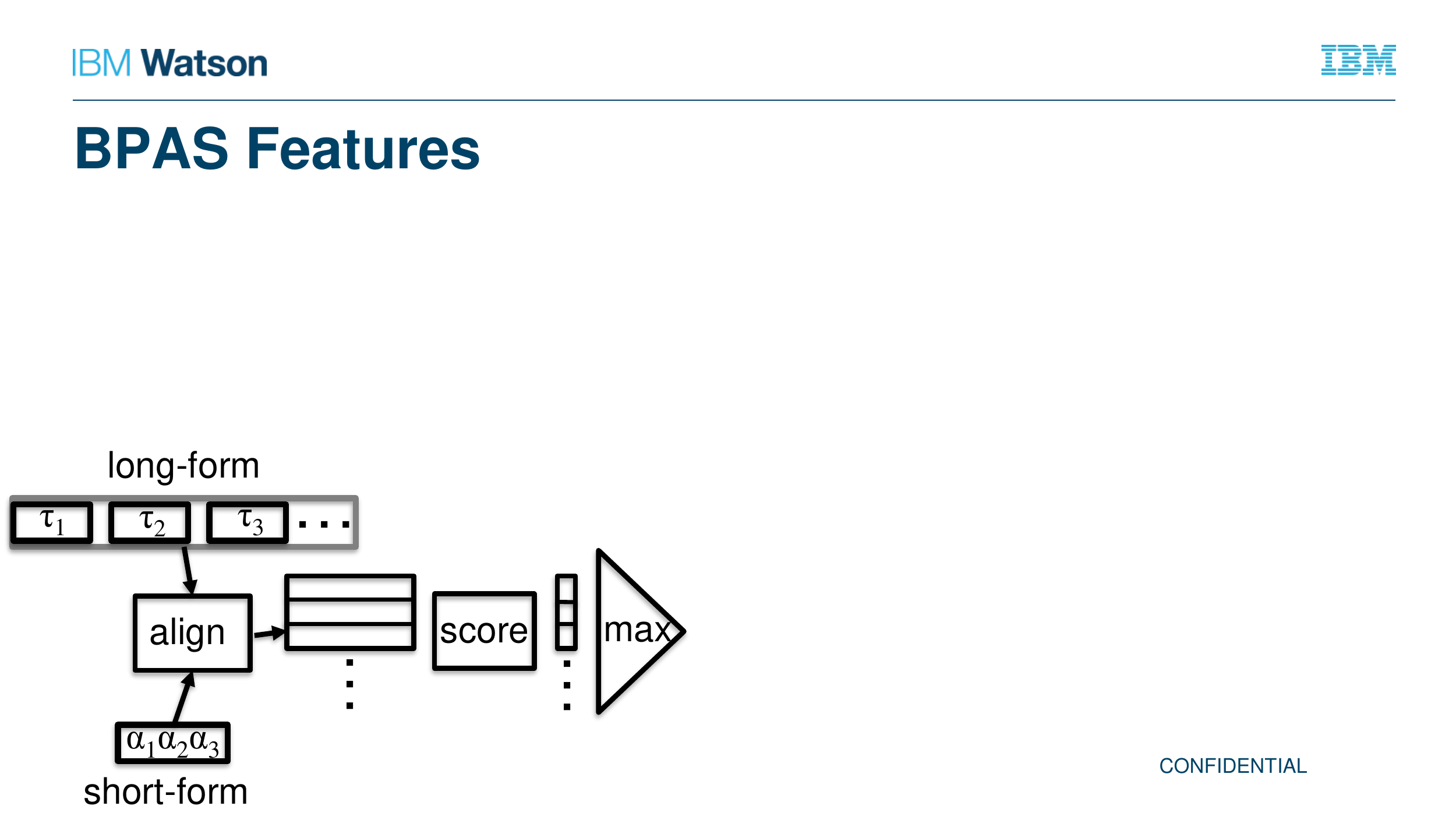}
   \caption{Maximum Over Alignments}
   \label{bpas}
  \end{figure}

After constructing the features of each alignment, a linear model scores the alignments. The surface similarity score is then the maximum over the alignment scores. The linear model to assess alignment quality is trained by backpropagation through the max pool. Figure \ref{bpas} illustrates the architecture of the surface similarity component.

The set of possible alignments is given by $A$ while the function to generate the features is $\gamma(\cdot) : A \mapsto \varmathbb{R}^{n}$ and the learned parameter vector is $\vec{\theta} \in \varmathbb{R}^{n}$. The surface similarity score $ss$ is then:
\begin{align*}
ss & = \max_{a \in A} \vec{\theta} \cdot \gamma(a)
\end{align*}

To reduce overfitting, we train this component separately, rather than end-to-end. The training maximizes the alignment model's individual contribution to scoring a candidate pair.

\section{Evaluation}

For each language, we split the short-forms into three folds for cross validation, training on two-thirds and testing on the remaining third.

We measure the precision and recall of the extracted short-form / long-form pairs by matching it with the ground truth constructed from Wikipedia redirects and disambiguation pages as described in Section \ref{sec:dataset}.
We report the area under the precision recall curve for each system on each language.
%TODO: rather pseudo-precision

In order to assess a more meaningful measure of recall, we filter the ground truth short-form / long-form pairs to include only those that co-occur within a document in the corpus used for abbreviation extraction. All existing methods of abbreviation extraction we reviewed develop the candidate set by finding the short-form and long-form in the same document, so this filtering step gives a more realistic upper-bound. With the filters applied the ground truth contains very few ($<$ 5\% by inspection) non-abbreviation pairs, meaning the measured recall for an extraction system is close to the true, theoretically achievable, recall. There is perhaps a bias in the ground truth towards containing more common or more important abbreviations, but this can be viewed as a point in the ground truth's favor. %The recall evaluated on our ground truth provides an unbiased estimate of an ``importance-weighted'' true recall.

Because the ground truth constructed by the method described in the Section \ref{sec:dataset} is certainly far from complete, we refer to the metric of precision obtained by judging true positives as only those that match this ground truth as the \textit{pseudo-precision}. We also ignore (rather than penalize) system pairs where the short-form is not present in the ground truth at all.

\section{Results}

Table \ref{resultsTable} gives the area under the (pseudo-)precision recall curve for each system on each of seven languages.  \textit{Candidate Scores} is a baseline system constructed by combining the method of Schwartz and Hearst \shortcite{Schwartz2003} as well as Candidate System 2 into a logistic model trained from their scores as features. The \textit{Word Embeddings} system augments that baseline with the semantic features described in the Semantic Similarity by Word Embeddings section. The \textit{Alignment} system further adds the alignment features from the Surface Similarity by Alignment section.

Scores in bold indicate a statistically significant improvement over the previous entry in the row with $p < 0.05$ by sampling permutation test. In the case of the \textit{Candidate Scores} system, bold indicates a statistically significant improvement over both previous scores.

The languages in Table \ref{resultsTable} are listed in order of the number of short-forms in the ground truth. For the languages with the least training data, the Alignment features do not produce significant improvement. This is perhaps because it is the most complex feature set.

Notably, the semantic features from word embeddings produced a large and significant improvement for all languages tested.

  \begin{figure}[hb]
   \centering
   \includegraphics[width=3.0in]{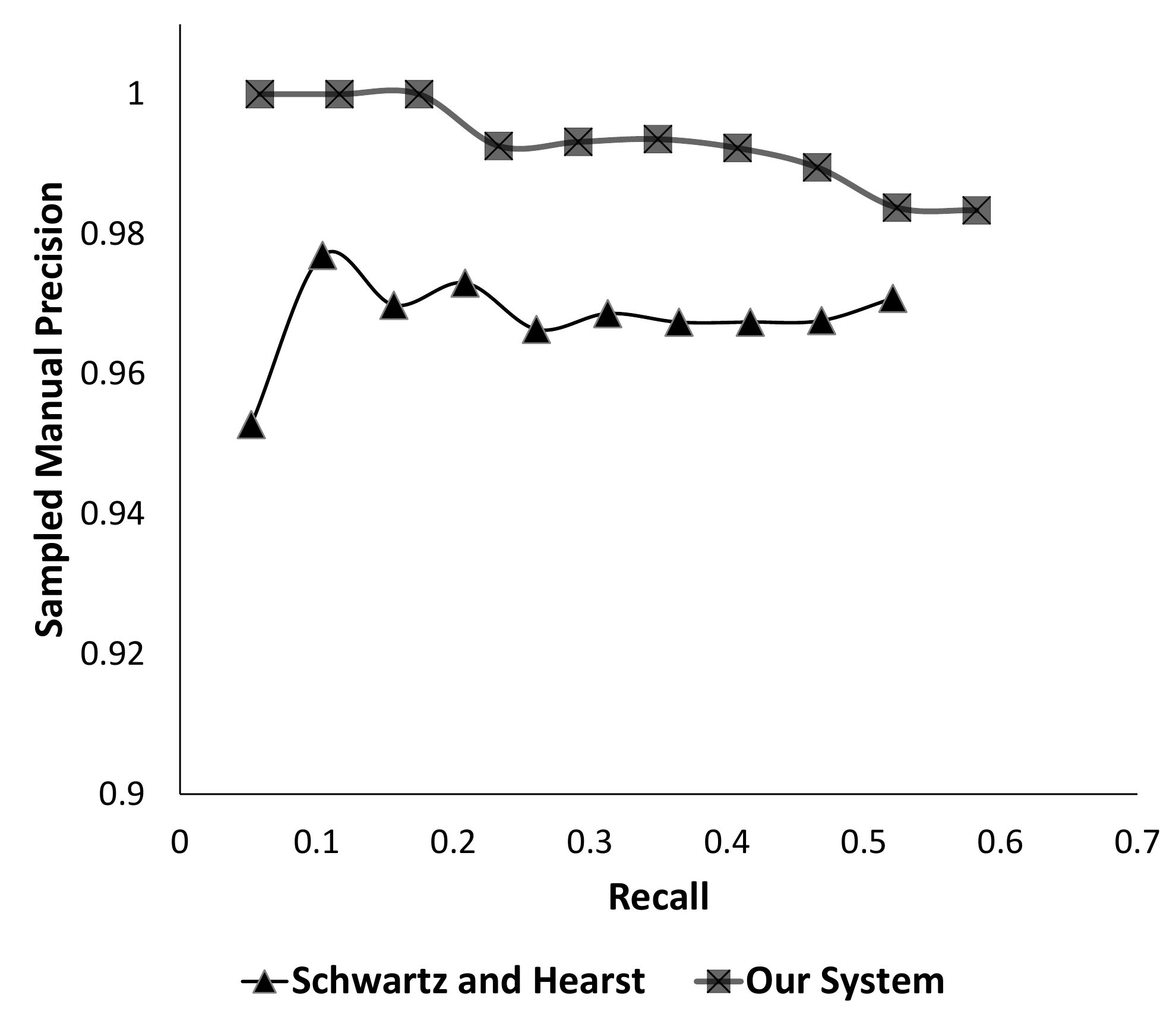} 
   \caption{English Sampled Precision Curves}
   \label{sampledPrecision}
  \end{figure}

\begin{table*}[b!]
\begin{center}
\begin{tabular}{|p{1.95cm}|p{1.95cm}|p{1.95cm}|p{1.95cm}|p{1.95cm}|p{1.95cm}|}
\hline \bf Language & \bf Schwartz and Hearst & \bf Candidate System 2 & \bf Candidate Scores & \bf Word \newline Embeddings & \bf Alignment \\ \hline \hline
English & 0.359 &	0.324 & \bf 0.408 & \bf 0.461 & \bf 0.480 \\ \hline 
German & 0.257 & 0.242 & \bf 0.329 & \bf 0.369 & \bf 0.437 \\ \hline 
French & 0.125 &	0.156 & \bf 0.222 & \bf 0.309 & \bf 0.335 \\ \hline
Italian & 0.273 & 0.304 & \bf 0.384 &	\bf 0.436 & \bf 0.473 \\ \hline 
Spanish & 0.381 & 0.350 & 0.432 & \bf 0.481 & 0.493 \\ \hline
Russian & 0.178 & 0.166 & \bf 0.253 & \bf 0.307 & 0.307 \\ \hline
Japanese & 0.259 & 0.025 & 0.259 & \bf 0.306 & 0.310 \\ \hline
\end{tabular}
\end{center}
\caption{\label{resultsTable} Areas Under Pseudo-Precision/Recall Curves}
\end{table*}

\begin{figure*}[hb]
   \centering
   \includegraphics[width=6.0in]{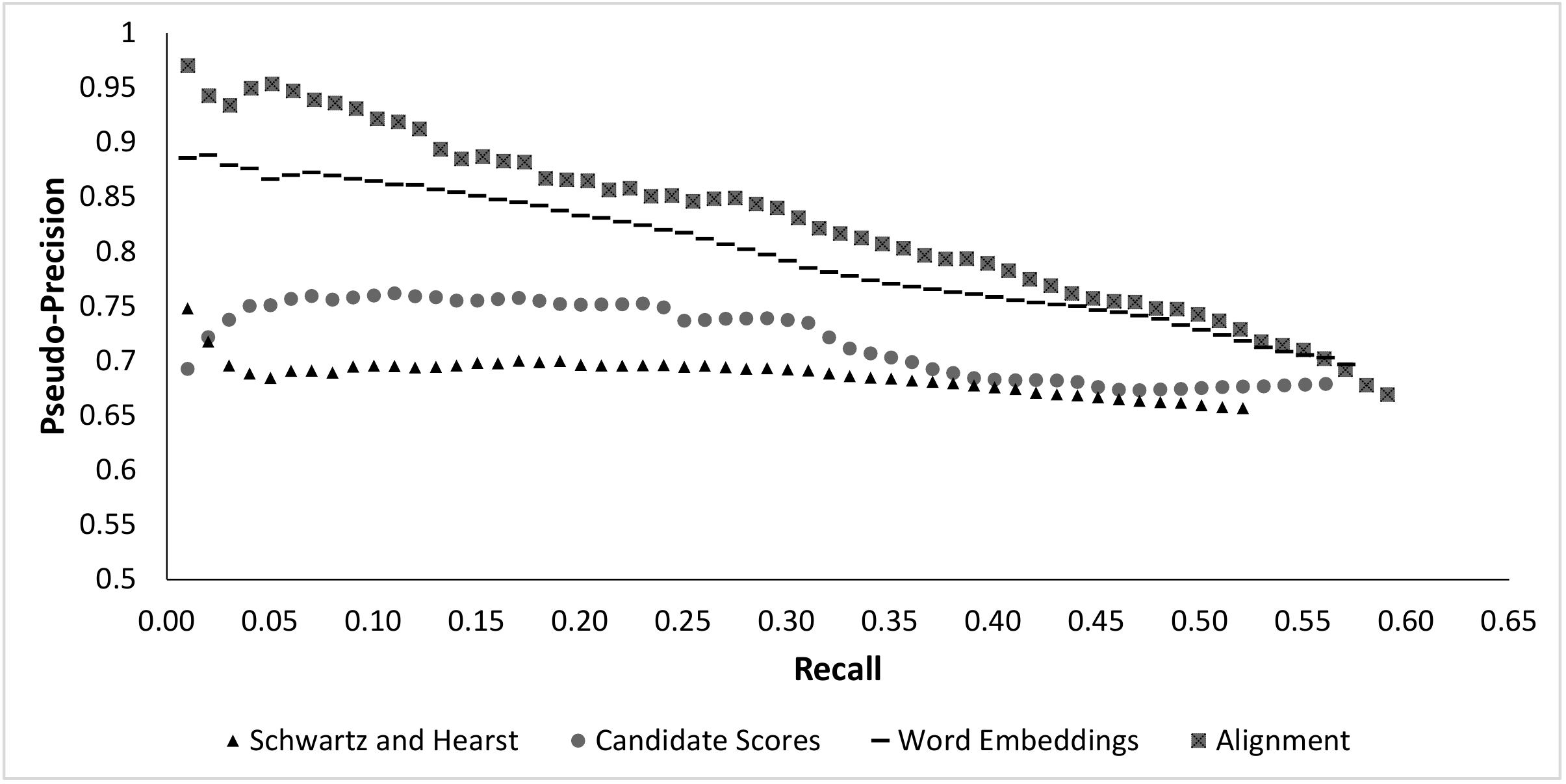} 
   \caption{English Pseudo-Precision Recall Curves}
   \label{enPseudoPR}
  \end{figure*}

\begin{figure*}[hb]
   \centering
   \includegraphics[width=6.0in]{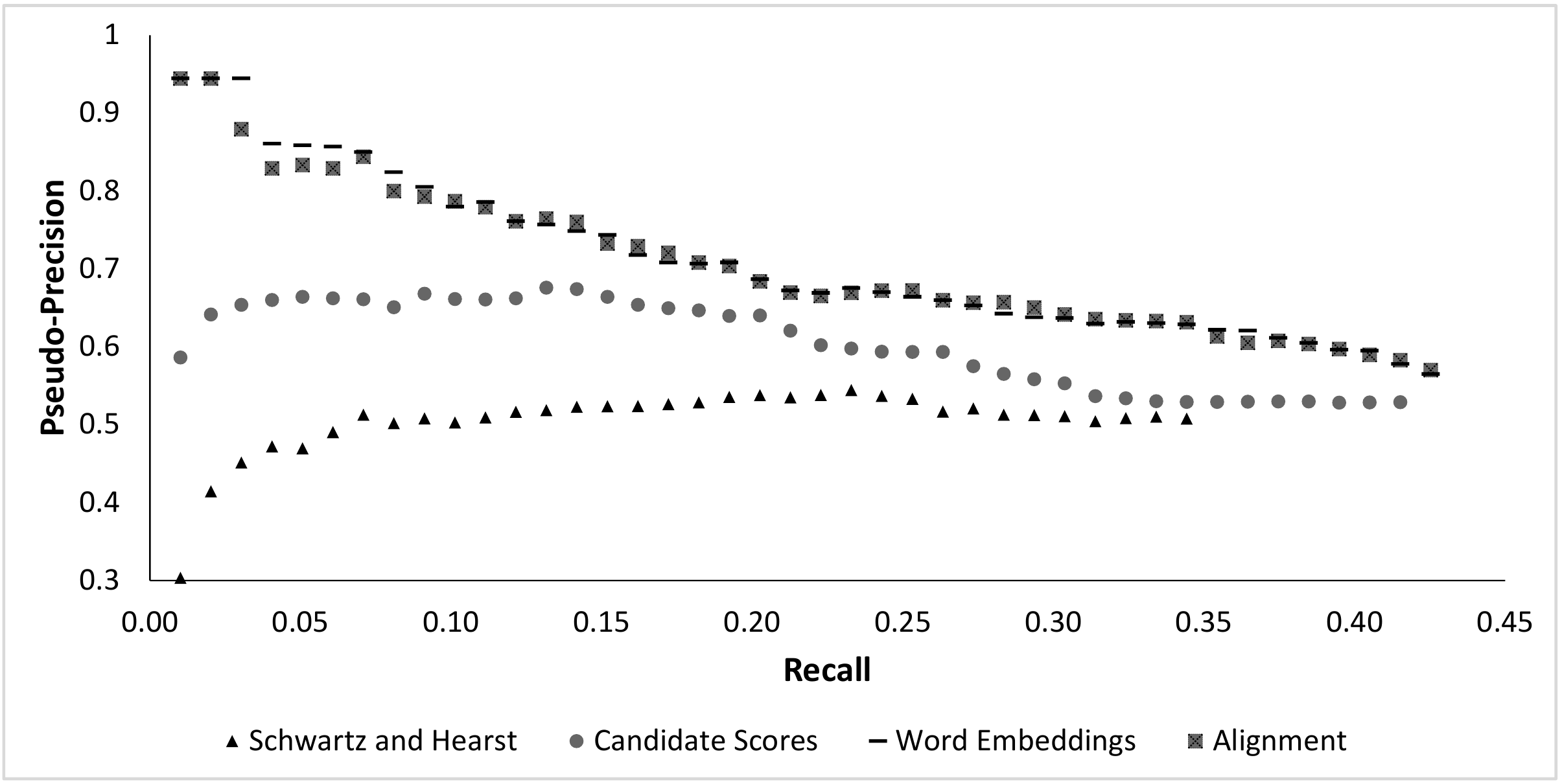}
   \caption{Russian Pseudo-Precision Recall Curves}
   \label{ruPseudoPR}
  \end{figure*}

%\begin{table}[htb!]
% % \centering
% \begin{tabular*}{\columnwidth}{@{\extracolsep{\stretch{1}}}*{7}{r}@{}}
%%\begin{tabular}{rll}
%\hline
%Method & AUC & F1 \\
%\hline
%NELL & 76.5\% &  67.3\% \\ 
%MLN & \textit{N/A} & \textit{N/A} \\
%PSL & 71.8\% & 80.4\% \\
%Consistency &  \textbf{89.5\%}  & \textbf{82.8\%} \\
%\end{tabular*}
%\caption{Performance on the full knowledge graph task}
%\label{tbl.fullTask}
%\end{table}

Figure \ref{enPseudoPR} shows the pseudo-precision and recall curves for the Schwartz and Hearst \shortcite{Schwartz2003} and Candidate Scores baselines, as well as the systems incorporating the semantic and semantic+alignment features. While Figure \ref{ruPseudoPR} gives the same results on Russian. As these examples illustrate, the missing area under the curve is mostly due to out-of-recall short-form / long-form pairs. Those that are found are fairly high (pseudo-)precision, even in the baseline systems.

\subsection{Sampled Manual Precision}
\label{secManualPrecision}

To quantify the under-estimate produced by pseudo-precision for our systems, we sample a number of extracted pairs at equally sized ranges of measured recall. The pairs sampled are drawn from those marked as false positives by the pseudo-precision metric. We then manually judge each pair as correct or incorrect. For each recall cutoff, the \textit{true error rate} ($ter$) is measured as:
\begin{align*}
ter & = (1 - \text{pseudo-precision}) \cdot \frac{\text{incorrect}}{\text{correct} + \text{incorrect}}
\end{align*}
The corrected precision is therefore $1-ter$.  By finding the precision for the extractions from each recall range, we produce a piece-wise constant approximation of the true precision-recall curve, reported as \textit{sampled manual precision}. 

Figure \ref{sampledPrecision} shows the sampled manual precision for English. Two-hundred randomly sampled pairs (from the output of each of the two systems graphed) judged as false positives by the incomplete ground truth were manually assessed, twenty from each of ten recall levels. As is clear, both systems are very accurate. This also suggests that much of the gain from the feature sets developed goes toward more highly scoring the dominant expansions of a short-form, the forms most likely to appear in the redirect or disambiguation pages.

%\FloatBarrier
\section{Conclusion}

We have created a large scale, multi-lingual, naturally current (through Wikipedia additions) resource for evaluating systems to mine abbreviations and their expansions. Using this resource we are able to apply a machine learning approach to incorporate the key aspects of the phenomenon of abbreviation: synonymy, topical relatedness and surface similarity. 

We show large, statistically significant improvements relative to strong rule-based baselines and a baseline that combines the rule-based systems into an ensemble. The contribution of semantic features is always large in the languages we tested while the contribution of a learned alignment model for surface similarity produces further gains when the available training data is large.

\clearpage
\bibliography{acronym}
\bibliographystyle{acl_natbib}

\end{document}